\documentclass[pdflatex,sn-mathphys-num]{sn-jnl}

\usepackage{graphicx}%
\usepackage{multirow}%
\usepackage{amsmath,amssymb,amsfonts}%
\usepackage{amsthm}%
\usepackage{mathrsfs}%
\usepackage[title]{appendix}%
\usepackage{xcolor}%
\usepackage{textcomp}%
\usepackage{manyfoot}%
\usepackage{booktabs}%
\usepackage{algorithm}%
\usepackage{algorithmicx}%
\usepackage{algpseudocode}%
\usepackage{listings}%
\usepackage{indentfirst} 
\usepackage{subcaption}
\usepackage{hyperref}

\theoremstyle{thmstyleone}%
\theoremstyle{thmstyletwo}%

\theoremstyle{thmstylethree}%

\begin{document}

\title[Article Title]{Comparative Strategies for Handling Imbalanced QSAR Regression: A Case Study on hERG Activity Prediction}


\author[1]{\fnm{Ming-Yen} \sur{Sung}}
\equalcont{These authors contributed equally to this work.}

\author[1,2]{\fnm{Hsin} \sur{Liu}}
\equalcont{These authors contributed equally to this work.}

\author[1]{\fnm{Tung-Sheng} \sur{Cho}}

\author[1]{\fnm{Cheng-Wei} \sur{Lee}}

\author*[2,3,4,5]{\fnm{Yufeng J.} \sur{Tseng}}\email{yjtseng@csie.ntu.edu.tw}

\affil[1]{%
  \orgname{VIRTUALMAN},
  \orgaddress{%
    \street{Fuxing Street},
    \city{Taipei},
    \postcode{106030},
    \country{Taiwan (ROC)}
  }
}

\affil[2]{%
  \orgdiv{Graduate Institute of Biomedical Engineering and Bioinformatics},
  \orgname{National Taiwan University},
  \orgaddress{%
    \street{No.~1, Sec.~4, Roosevelt Road},
    \city{Taipei},
    \postcode{10617},
    \country{Taiwan (ROC)}
  }
}

\affil[3]{%
  \orgdiv{Department of Computer Science and Information Engineering},
  \orgname{National Taiwan University},
  \orgaddress{%
    \street{No.~1, Sec.~4, Roosevelt Road},
    \city{Taipei},
    \postcode{10617},
    \country{Taiwan (ROC)}
  }
}

\affil[4]{%
  \orgdiv{The Metabolomics Core Laboratory, Center of Genomic Medicine},
  \orgname{National Taiwan University},
  \orgaddress{%
    \street{No.~1, Sec.~1, Jen Ai Road},
    \city{Taipei},
    \postcode{10617},
    \country{Taiwan (ROC)}
  }
}

\affil[5]{%
  \orgdiv{Drug Research Center, School of Pharmacy, College of Medicine},
  \orgname{National Taiwan University},
  \orgaddress{%
    \street{No.~1, Sec.~4, Roosevelt Road},
    \city{Taipei},
    \postcode{10617},
    \country{Taiwan (ROC)}
  }
}


\abstract{Data imbalance frequently occurs in QSAR modeling, yet most studies focus on classification tasks, and few address regression tasks, as it is challenging to determine which regions of the data are sparse or dense. Data-level and algorithm-level imbalance-handling strategies can have different effects on model performance. In this study, we systematically compare various strategies across GCN and fingerprint-based machine learning models. To evaluate model performance more meaningfully, we partitioned the target region into several subregions based on both domain knowledge and data distribution. We focused on the region of greatest practical interest. Our results demonstrate that selecting models based on this interest-region metric leads to more reliable performance than using the total-region metric. We also observed that different imbalance-handling strategies can have divergent effects depending on the model type. These findings highlight the importance of region-aware evaluation in regression QSAR modeling and provide practical guidance for selecting appropriate strategies for imbalanced datasets.}




\maketitle

\section*{Scientific Contribution}

This study uses hERG activity prediction as a case study to systematically evaluate KDE-based reweighting and resampling strategies for imbalanced regression tasks, a common challenge in QSAR modeling. Rather than relying solely on conventional whole-region metrics, we propose a region-aware evaluation and model selection framework that integrates data distribution characteristics with biological domain knowledge to prioritize practically important target regions. Furthermore, we demonstrate that fingerprint-based and graph-based models exhibit distinct responses to imbalance-handling strategies, emphasizing the need for model-specific optimization in regression QSAR tasks.

\section{Introduction}\label{sec1}

Quantitative structure--activity relationship (QSAR) modeling is an essential technique in cheminformatics and computational drug discovery \cite{liReviewQuantitativeStructureactivity2025}. Depending on how molecular features are extracted, QSAR models can be broadly classified as fingerprint-based or graph-based \cite{vasilevComprehensiveReviewApplication2025}. These models establish mathematical relationships between molecular structures and their biological or physicochemical properties, enabling applications in potency optimization, toxicological risk assessment, and environmental chemistry. Regardless of the model type, QSAR models often encounter data imbalance \cite{zakharovQSARModelingImbalanced2014} \cite{lovricPredictiveCapabilityQSAR2021}, particularly in toxicity prediction tasks \cite{liImbalancedToxicityPrediction2020} \cite{al-jubouriImbalanceRegressionApproach2025}. Imbalanced problems have two characteristic features: an uneven distribution of the target variable and a particular emphasis on sparse but practically important cases \cite{ribeiroImbalancedRegressionExtreme2020}. For example, in many experimental settings, only a small proportion of compounds exhibit toxic responses, which are the samples of primary interest.

Most efforts focus on classification, while relatively few address regression, because the concept of ``minority'' or ``rare'' samples is more difficult to define in continuous outcomes \cite{ribeiroImbalancedRegressionExtreme2020}. This issue not only affects training--such as which regions should be oversampled or assigned higher weights---but also has important implications for model evaluation and selection. Regression approaches can generally be categorized into algorithm-level and data-level strategies \cite{steiningerDensitybasedWeightingImbalanced2021}. Algorithm-level methods assign different weights to samples, thereby modifying the gradient during model training. In contrast, data-level methods alter the sample distribution--e.g., by downsampling dense regions--to improve model performance. Many studies have evaluated model performance by dividing the target region into multiple regions rather than relying solely on total MSE \cite{yangDelvingDeepImbalanced2021} \cite{zhangNovelGradientBoosting2024}, highlighting the importance of region-aware assessment.

However, several limitations remain. First, in QSAR research, it is rarely discussed which imbalance-handling strategies are most suitable for different model types (e.g., graph-based vs. fingerprint-based models). Second, commonly used bin definitions often lack biological or experimental interpretability, making it unclear whether improvements in a given region are meaningful in practice. 

To address these gaps, this study investigates imbalance-handling strategies in graph-based and fingerprint-based models. We introduce a region definition scheme that combines domain knowledge with kernel density estimation (KDE) \cite{steiningerDensitybasedWeightingImbalanced2021}, enabling biologically meaningful evaluation of model performance. Using a human Ether-à-go-go-Related Gene (hERG) cardiotoxicity assay dataset as an example, we examine how reweighting, downsampling, and Synthetic Minority Oversampling for Regression (SMOGN) \cite{brancoSMOGNPreprocessingApproach2017} strategies affect model performance within these regions. Model evaluation and selection are conducted in a region-aware manner to focus on the response intervals that are most practically important.

The contributions of this paper are as follows:
\begin{enumerate}
    \item We propose a region definition framework that integrates data-driven density estimation with domain knowledge for biologically meaningful assessment in toxicity prediction.
    \item We systematically compare algorithm-level (reweighting) and data-level (downsampling, SMOGN) strategies across both graph-based and fingerprint-based models.
    \item We adapt SMOGN by manually defining sparse and dense regions, leveraging KDE-based density estimation and experimental knowledge, to improve model performance in toxicity prediction.
    \item We demonstrate that region-aware evaluation and selection reveal performance patterns that global metrics would obscure, guiding the development of the QSAR model in imbalanced regression settings.
\end{enumerate}

\section{Methods}\label{sec2}

\subsection{Dataset}\label{subsec1}

The dataset used in this study was obtained from the public PubChem database (AID: 588834) \cite{AID588834QHTS}. It is the result of the FluxOR\textsuperscript{\texttrademark} thallium flux assay \cite{FluxORPotassiumIon}, which is commonly employed for preliminary screening of potential hERG-risk compounds. The dataset includes a total of 5,381 compounds, some of which appear multiple times because they originate from different sources. Each compound contains concentration--response curve (CRC) information, which is often converted into IC\textsubscript{50} values. In this study, however, we directly use the activity at different concentrations to train the model. The concentration range spans from 0.0000049913~\(\mu\)M to 38,440.0~\(\mu\)M. Activity is expressed as a percentage, which is calculated from negative and positive controls (negative control = 0\%, positive control = \(-100\%\)) \cite{krishnaHighThroughputChemicalScreening2022}. Consequently, negative values are the primary focus, although positive values are also present. All measurements are included in the model to allow comprehensive learning of both risk and non-risk structures.

\subsection{Splits preparation}\label{subsec2}

\subsubsection{Split by compound}\label{subsubsec1}
To enable the model to predict unseen structures, the dataset was split into training, validation, and test sets in an 8:1:1 ratio per compound, using different random seeds. The split was performed using the \texttt{train\_test\_split} function from scikit-learn (version 1.3.0).

\subsubsection{Features generation}\label{subsubsec2}
Fingerprints are vector representations of compounds that encode their structural features, such as the presence, counts, or degree of certain substructures \cite{liFingerprintingInteractionsProteins2024}. In this study, we utilized Extended-Connectivity Fingerprints (ECFP), a type of circular topological fingerprint. An ECFP is generated by iteratively assigning atom identifiers within increasing radii of perception, up to a maximum radius \(n\). Through this process, all substructures originating from different atoms are systematically captured and represented as vectors \cite{rogersExtendedConnectivityFingerprints2010}. The length of the fingerprint also plays a crucial role: longer fingerprints may retain more structural information but incur higher computational costs during model training. ECFPs were generated based on SMILES using the \texttt{rdFingerprintGenerator.GetMorganGenerator} function in RDKit (version 2025.03.5), with default parameters (radius = 3 and fpSize = 2048) and expressed in ECFP6.

For graph-based models, the compound input graphs were generated from SMILES using DeepChem (version 2.7.1) and RDKit. The features are a weighted combination of the inputs, and the weights are updated with the training model at the same time.

\subsubsection{Converting CRC into activity at one concentration}\label{subsubsec3}
To augment the training set, each CRC series was decomposed into multiple individual data points, each representing the activity at a specific concentration. This approach increases the number of training samples and allows the model to predict activity at any concentration within the tested range.

Averaging was applied to the training, validation, and test sets. Each experiment was tuned using different random seeds for the training and validation sets, and performance was evaluated on the corresponding test set.

\subsection{Kernel density estimation}\label{subsec3}
Kernel Density Estimation (KDE) is a non-parametric statistical method used to estimate the Probability Density Function (PDF) of a random variable. Unlike histograms, which rely on discrete bins, KDE uses a kernel function to produce a smooth, continuous curve that represents the data distribution. In this article, we use the \texttt{gaussian\_kde} function from the SciPy library (version 1.7.3) to generate the density estimate. The bandwidth ($bw$), which controls the smoothness of the density curve, was selected based on standard rules (Scott's and Silverman's) as well as fixed scalar constants (0.05, 1.0).

\subsection{Imbalance regression approaches}\label{subsec4}
To address the imbalance in regression tasks, we employed KDE-based strategies for reweighting, downsampling, and SMOGN. For reweighting, sample weights were assigned as the reciprocals of the KDE-estimated density, then shifted so that the minimum weight equals 1. For downsampling, the KDE was used to define dense and sparse regions; samples in the dense region were selectively reduced, while samples in the sparse region were retained. For SMOGN, which combines downsampling of dense regions and oversampling of sparse regions, we used the KDE results to manually define the control points that delineate the dense and sparse regions. By default, SMOGN automatically determines the control points; however, the resulting performance was suboptimal. Therefore, we manually defined the control points based on KDE-guided distribution analysis and further adjusted the downsampling ratio to improve model performance.

\subsection{ML model development}\label{subsec5}
We applied two different ensemble tree algorithms to develop machine learning models.

Random Forest (RF) introduces two sources of randomness to reduce the variance of individual tree estimators, ensuring that each tree differs from the others. Each tree is trained independently, and the final prediction is obtained by aggregating the results of all trees via bagging (voting), which helps mitigate overfitting. RF models were implemented using the \texttt{RandomForestRegressor} class from scikit-learn (version 1.0.2).

In contrast, all trees in XGBoost (XGB) are dependent. The model defines an objective function that includes both the training loss and a regularization term, and sequentially builds trees to optimize it. Each tree is constructed based on the performance of the previous trees to minimize the loss. Early stopping can be applied in XGB to prevent overfitting. XGB models were implemented using the \texttt{XGBRegressor} class (version 1.6.2).

To identify the optimal hyperparameter combination, the \texttt{RandomizedSearchCV} function from scikit-learn was used. The ranges of hyperparameters explored for both RF and XGB are listed in Table~\ref{tab:hyperparameter_ranges}. To ensure sufficient tree growth in XGB, the number of estimators (\texttt{n\_estimators}) was set to 1000, and early stopping (\texttt{early\_stopping\_rounds}) was set to 30 to prevent overfitting.

\begin{table}[ht]
\centering
\caption{Hyperparameter ranges for Random Forest (RF) and XGBoost (XGB) models.}
\label{tab:hyperparameter_ranges}
\begin{tabular}{lll}
\toprule
Algorithm & Hyperparameter & Range \\
\midrule
\multirow{4}{*}{RF} 
& max\_features & 0.1, 0.2, 0.4, 0.8, 1 \\
& max\_depth & 5, 10, 15, 20, 25, None \\
& max\_leaf\_nodes & 10, 100, 200, None \\
& min\_samples\_leaf & 1, 2, 5, 10, 20, 50 \\
\midrule
\multirow{5}{*}{XGB} 
& learning\_rate & 0.001, 0.01, 0.1, 0.3 \\
& gamma & 0, 1, 5, 10 \\
& min\_child\_weight & 1, 5, 10, 20 \\
& colsample\_bytree & 0.4, 0.6, 0.8, 1 \\
& max\_depth & 3, 6, 9, 12 \\
& subsample & 0.5, 0.7, 1 \\
\bottomrule
\end{tabular}
\end{table}

\subsection{GCN model development }\label{subsec6}
Compound information was generated from SMILES using DeepChem  and RDKit. The GCN model takes these inputs, represented as features \cite{deepchemGCNTutorial}, and optimizes them during training. The GCN architecture consists of two \texttt{SparseGraphConvSimple} layers, each followed by a \texttt{BatchNormalization} layer, and six fully connected (dense) layers with a dropout rate of 0.1. The learning rate was set to 0.0001. The MSE was employed as a loss function during training.

\subsection{Region-aware model evaluation and model selection}\label{subsec7}

In addition to the imbalance-handling strategies applied during model training, evaluating performance across the entire target range is insufficient for toxicity prediction tasks. Previous studies typically partitioned the target space into many-shot, medium-shot, and few-shot regions solely based on sample size \cite{yangDelvingDeepImbalanced2021}, a partitioning that lacks direct experimental and biological interpretation.

In this study, target regions were defined according to the experimental characteristics of the FluxOR\texttrademark{} thallium flux assay. Specifically, the response range was partitioned into five intervals: $[-200, -100)$, $[-100, -20)$, $[-20, 20)$, $[20, 100)$, and $[100, 200)$. Prior studies have reported that approximately 20\% hERG channel block can lead to QTc interval prolongation \cite{kaczorowskiImportanceBeingProfiled2011}, motivating the selection of $-20$ as a critical threshold. In addition, $-100$ corresponds to the positive control in the FluxOR\texttrademark{} assay. The interval $[-200, -100)$ was retained as a single region due to its limited sample size, which precluded further subdivision. The positive response range was partitioned symmetrically with respect to the negative range to maintain consistency in region definition. Based on these biologically and experimentally meaningful thresholds, region-specific performance metrics were reported.

Among these regions, the interval $[-100, -20)$ is of particular interest, as it represents moderate hERG inhibition levels that are pharmacologically relevant and challenging to predict. Therefore, model selection was primarily based on performance within this region.

\section{Results}\label{sec3}

\subsection{Dataset distribution characteristics and density interval segmentation}
\begin{figure}[h]
    \centering
    \includegraphics[width=0.9\textwidth]{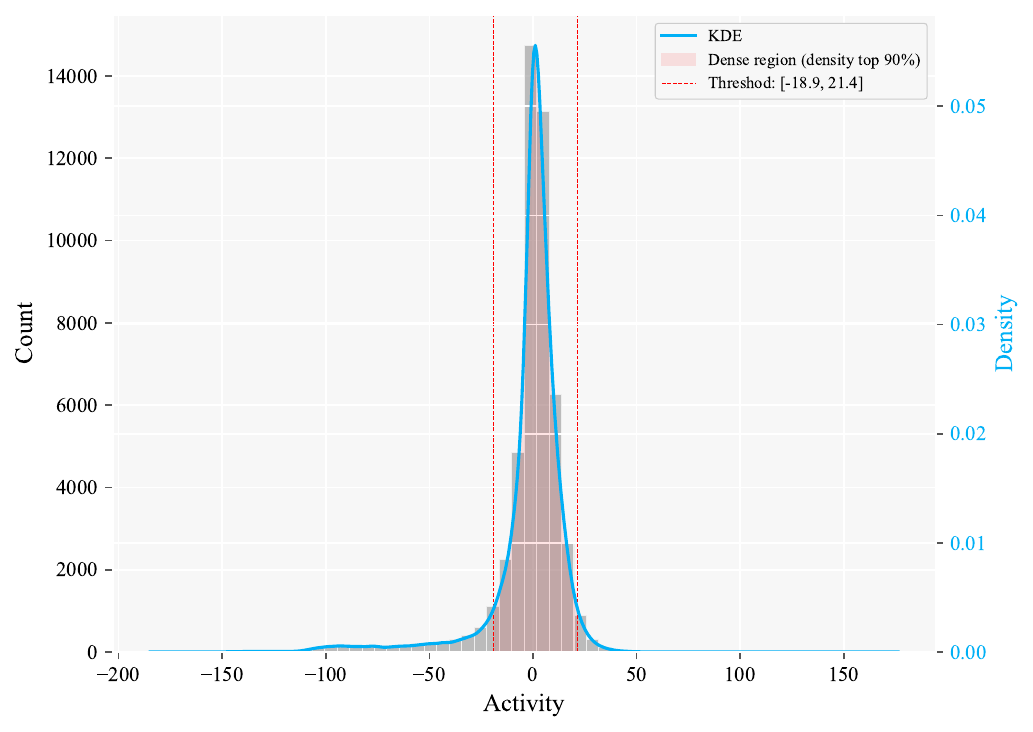}
    \caption{Identification of the highest density region (HDR) for activity distribution using KDE}
    \label{dense_region_illustration}
\end{figure}

As shown in Supplementary Figure 1, as the bandwidth factor increases, the resulting KDE curve becomes excessively smooth, making subtle differences or local variations in the data distribution less pronounced. We employed the default Scott's rule for bandwidth estimation as it yielded a KDE curve that best reflected the structure observed in the empirical histogram. Based on the resulting density estimate (Figure~\ref{dense_region_illustration}), the 90\% Highest Density Interval (HDI) spans from $-18.9$ to $21.4$. In other words, only 10\% of the data fall outside this interval, highlighting a pronounced imbalance in the data for the regression task. Importantly, prior studies have shown that hERG-related toxicity typically begins to manifest around $-20$ \cite{kaczorowskiImportanceBeingProfiled2011}, with increasingly negative values associated with elevated toxicity risk. For this reason, we designate the region within the HDI ($-18.9$ to $21.4$) as the dense region, corresponding to minimal hERG effect. Values outside this interval constitute the sparse region, with the negative sparse region (less than $-18.9$) being particularly relevant because of its strong association with hERG toxicity. With this characterization of dense and sparse regions, we proceeded to evaluate several strategies to address the imbalance in this regression problem.

\subsection{Different imbalance strategies applied to training}
\begin{figure}[h]
    \centering
    \includegraphics[width=0.9\textwidth]{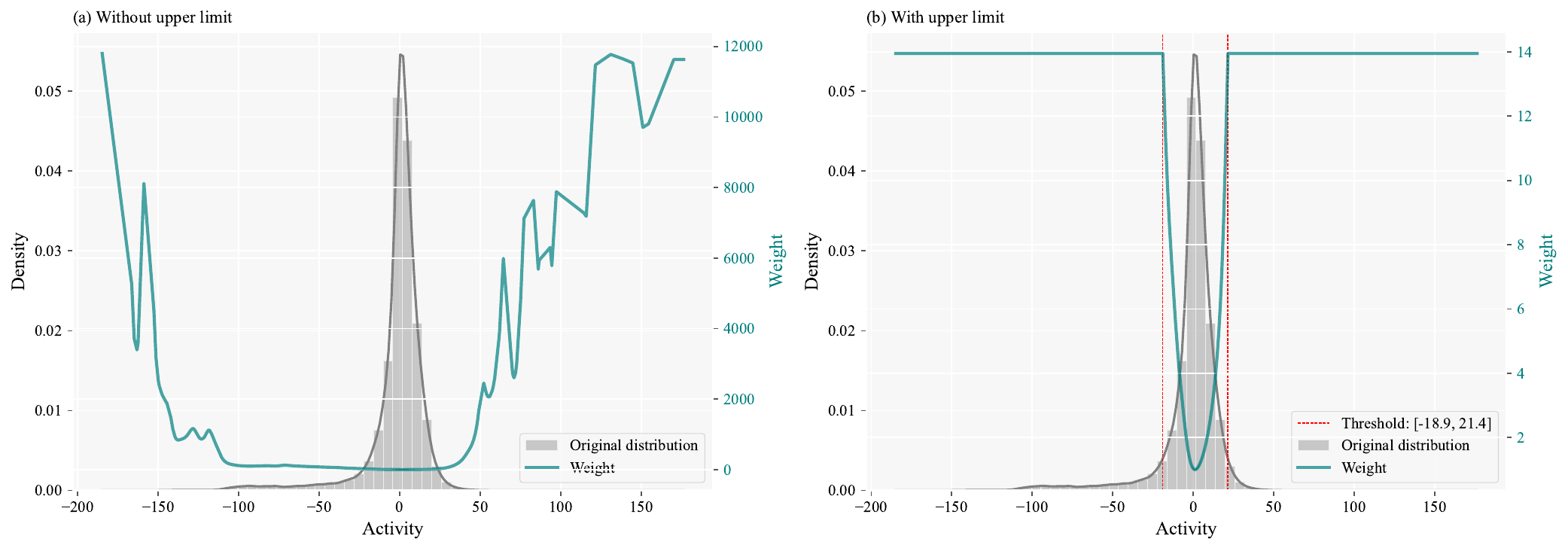}
    \caption{KDE-based reweighting}
    \label{reweighting}
\end{figure}

The KDE-based weight distribution is shown in Figure~\ref{reweighting} (a). The weights range from 1 to nearly 12,000, which could cause the model to overemphasize the sparse region, leading to training instability. To address this issue, we imposed an upper bound on the weights. Specifically, a 90\% density threshold was applied as the cutoff. All samples in the sparse region were assigned the same capped weight, while samples in the dense region retained their relative weight differences based on the KDE values (Figure~\ref{reweighting} (b)) . This approach ensures model stability while preserving meaningful distinctions within the dense region. 

\begin{figure}[h]
    \centering
    \includegraphics[width=0.9\textwidth]{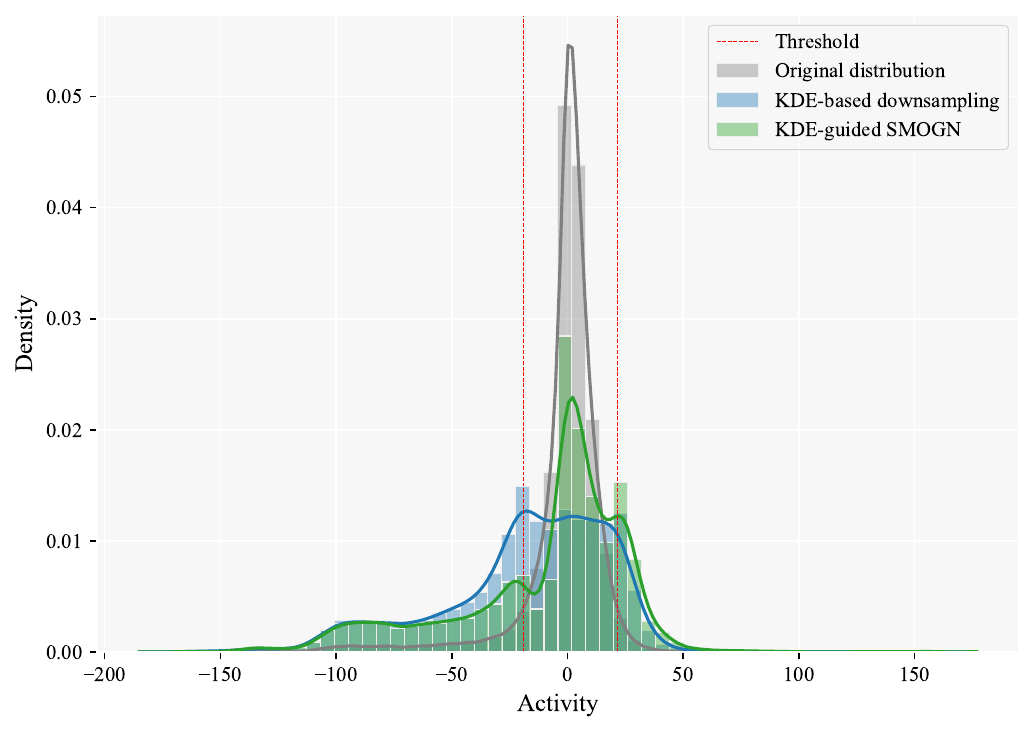}
    \caption{Comparison of Resampling Strategies: Original Distribution vs. KDE-Downsampling and SMOGN}
    \label{kde_resampling}
\end{figure}

Figure~\ref{kde_resampling} shows the data distributions for KDE-based downsampling, KDE-based SMOGN, and the original dataset. Both KDE-based methods mitigate the imbalance between dense and sparse regions, creating a more uniform distribution that supports stable model training.

\clearpage

\subsection{Model selection based on total MSE}
\begin{figure}[h]
    \centering
    \includegraphics[width=0.9\textwidth]{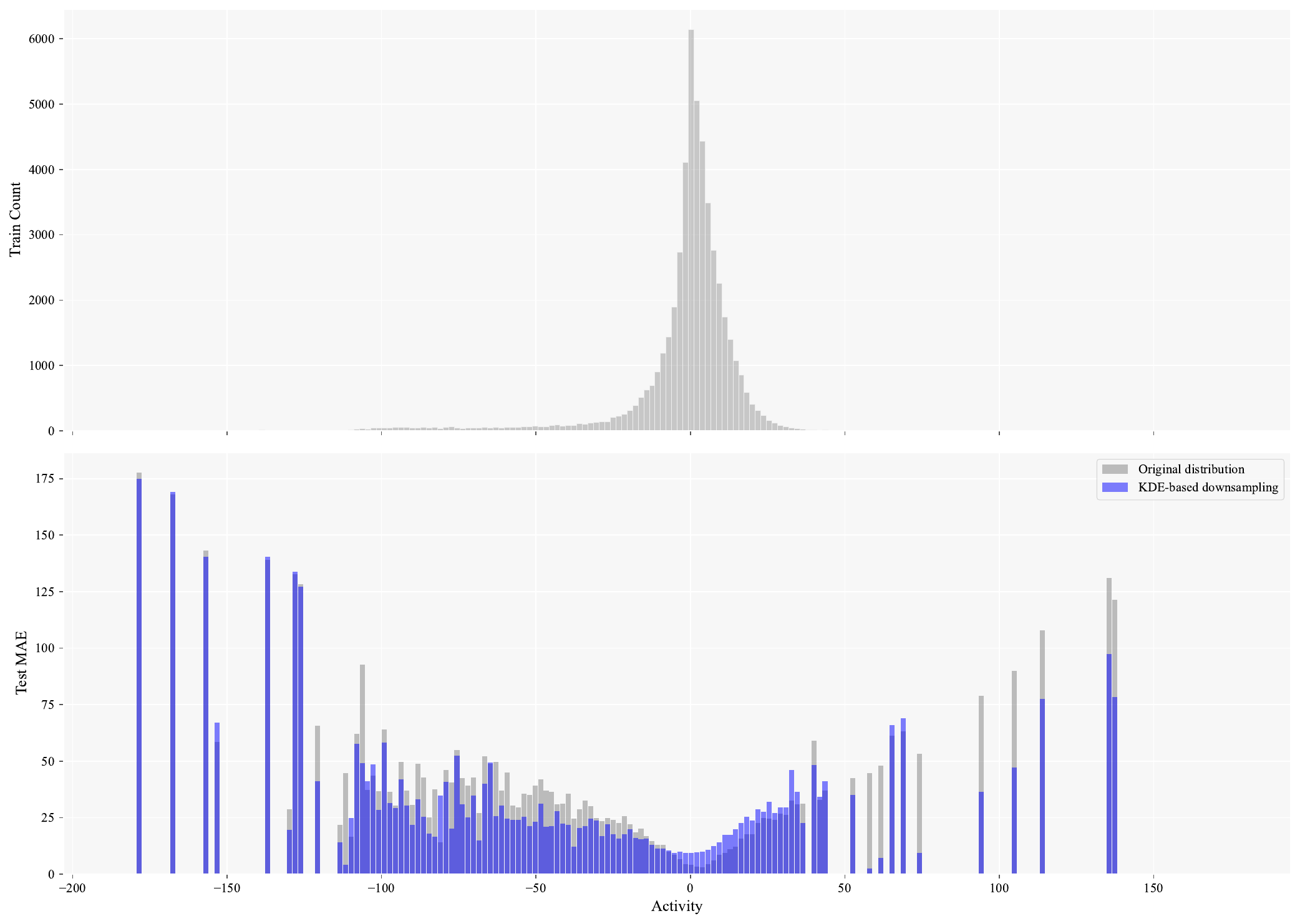}
    \caption{Relationship between training data distribution and test MAE for GCN under the downsampling strategy.}
    \label{test_MAE_undersampling}
\end{figure}

We compared the effects of three strategies across different response regions using GCN, RFR, and XGBR models. Based on Table~\ref{tab:tst_total_mse} and Figure~\ref{test_MAE_undersampling}, the best model is selected by minimizing the mean squared error (MSE) over the entire target range. Under this selection criterion, none of the three imbalance-handling strategies leads to a noticeable improvement in overall performance. This outcome is expected, as these strategies are applied only to the training data, while the validation and test sets preserve the original, highly imbalanced target distribution. Consequently, a global error metric is dominated by dense regions and remains insensitive to improvements in sparse regions.

This observation highlights that addressing data imbalance in regression requires not only data-level interventions during training, but also appropriate model selection criteria and evaluation metrics that explicitly account for the importance of sparse regions.

\begin{table}
\caption{Test performance of different imbalance-handling strategies based on total MSE selection.}
\label{tab:tst_total_mse}
\begin{tabular}{llllll}
\toprule
Model & Treatment & Bin & MAE & Pearson & Spearman \\
\midrule
GCN & Original & total & 9.74 ± 0.5 & 0.55 ± 0.07 & 0.23 ± 0.02 \\
GCN & KDE-based reweighting & total & 11.87 ± 0.83 & 0.51 ± 0.05 & 0.2 ± 0.01 \\
GCN & KDE-guided SMOGN & total & 17.75 ± 1.73 & 0.42 ± 0.02 & 0.22 ± 0.03 \\
GCN & KDE-based downsampling & total & 14.25 ± 0.8 & 0.45 ± 0.03 & 0.2 ± 0.02 \\
RFR & Original & total & 9.57 ± 0.19 & 0.57 ± 0.02 & 0.26 ± 0.01 \\
RFR & KDE-based reweighting & total & 10.39 ± 0.38 & 0.57 ± 0.02 & 0.25 ± 0.01 \\
RFR & KDE-guided SMOGN & total & 15.18 ± 0.53 & 0.18 ± 0.17 & 0.05 ± 0.08 \\
RFR & KDE-based downsampling & total & 12.76 ± 0.31 & 0.52 ± 0.03 & 0.23 ± 0.01 \\
XGBR & Original & total & 9.55 ± 0.26 & 0.58 ± 0.03 & 0.26 ± 0.02 \\
XGBR & KDE-based reweighting & total & 10.58 ± 0.4 & 0.55 ± 0.02 & 0.24 ± 0.01 \\
XGBR & KDE-guided SMOGN & total & 11.1 ± 0.2 & 0.41 ± 0.03 & 0.18 ± 0.01 \\
XGBR & KDE-based downsampling & total & 11.01 ± 0.13 & 0.5 ± 0.02 & 0.23 ± 0.02 \\
\bottomrule
\end{tabular}
\end{table}

\subsection{Performance breakdown across target regions}
To address the limitations of using total-region validation MSE as a metric, the response space was divided into several intervals, as listed in Supplementary Table 1, 2, and 3, based on sample size and biological relevance. In particular, $-20$ marks the onset of hERG activity, while $-100$ corresponds to the reference activity of the positive control in the assay. The region $[-200,-100)$ contains only 42 samples, making further subdivision impractical; it also cannot be combined with $[-100,-20)$ without exceeding the biologically meaningful reference. For symmetry, the positive range was divided into equal intervals. The interval $[-20,20)$ coincides with the dense region threshold, containing the majority of samples and therefore dominating global metrics.

As shown in Supplementary Table 1 (GCN), the interval $[-20,20)$ contributes the most to the overall metric due to the highly imbalanced data distribution. Notably, all three strategies consistently improve predictive performance in the $[-100,-20)$ region, which is of primary interest in hERG studies. However, performance in the remaining regions shows little or no improvement, suggesting a potential trade-off between optimizing sparse, high-interest regions and maintaining performance elsewhere.

For the RFR model, only SMOGN fails to improve performance in the $[-20,20)$ region, not only in MAE but also in Pearson and Spearman correlations. In contrast, none of the evaluated strategies consistently improve performance in the XGBR model. The underlying reasons for these model-dependent differences warrant further investigation.

In summary, beyond the choice of preprocessing strategies used during model training, model evaluation should prioritize metrics computed over application-relevant response regions (i.e., the $[-100,-20)$ region in this study), rather than relying solely on global performance metrics dominated by densely populated regions.

\subsection{Model selection based on interest-aware criteria}
\begin{figure}[h]
    \centering
    \includegraphics[width=0.9\textwidth]{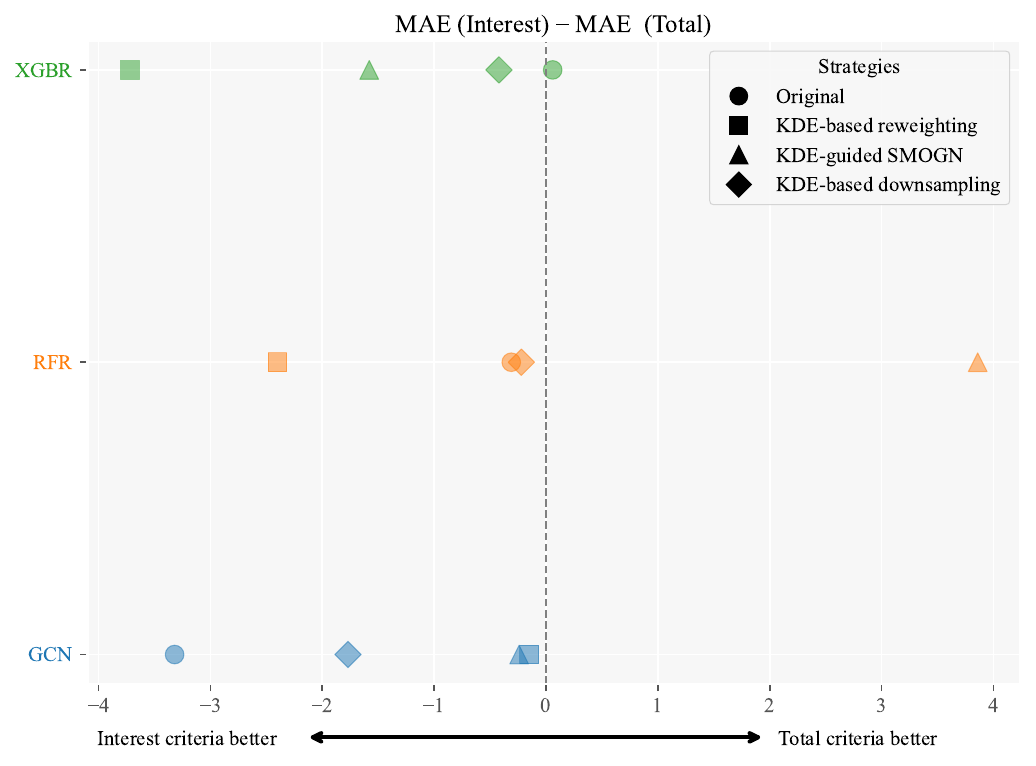}
    \caption{The impact of model selection strategy on test MAE.}
    \label{different_strategies}
\end{figure}

As discussed above, imbalance-aware strategies primarily improve predictive performance in the $[-100,-20)$ region. Accordingly, this section focuses on model selection based on region-specific performance. Instead of selecting models based on the global validation MSE (valMSE), we use the valMSE computed over the $[-100,-20)$ region as the model selection criterion. The corresponding results are summarized in Figure~\ref{different_strategies} and Table~\ref{tab:model_selection_interest_region}.

Overall, models selected based on the $[-100,-20)$ valMSE consistently achieve better performance in this target region compared to those selected using the global valMSE, indicating that interest-aware model selection is effective.

When focusing on performance in the $[-100,-20)$ region, KDE-based reweighting yields more pronounced improvements for RFR and XGBR models, whereas its effect on GCN remains limited. In contrast, downsampling strategies substantially improve performance in GCN and RFR, but show minimal benefit for XGBR, highlighting clear model-dependent differences.

Across all settings, improvements are primarily reflected in MAE, whereas correlation metrics remain largely unchanged. Notably, although SMOGN yields substantial MAE improvements in GCN, it comes at the cost of reduced correlation performance. Furthermore, SMOGN degrades performance in both RFR and XGBR models, suggesting that its effectiveness is highly sensitive to model architecture.

\begin{sidewaystable}[ht]
\centering
\caption{Model selection effect on [-100,-20) region performance}
\label{tab:model_selection_interest_region}
\begin{tabular}{llllllll}
\toprule
\multicolumn{2}{c}{} & \multicolumn{3}{c}{Interest\_Criteria} & \multicolumn{3}{c}{Total\_Criteria} \\
model & Treatment & tst\_mae & tst\_pearson & tst\_spearman & tst\_mae & tst\_pearson & tst\_spearman \\
\midrule
GCN & Original & 29.56 ± 1.37 & 0.56 ± 0.03 & 0.53 ± 0.04 & 32.88 ± 2.43 & 0.54 ± 0.05 & 0.51 ± 0.04 \\
GCN & KDE-based reweighting & 29.01 ± 2.51 & 0.54 ± 0.05 & 0.51 ± 0.04 & 29.16 ± 0.59 & 0.55 ± 0.04 & 0.52 ± 0.03 \\
GCN & KDE-guided SMOGN & 23.38 ± 1.53 & 0.49 ± 0.04 & 0.49 ± 0.04 & 23.62 ± 1.54 & 0.5 ± 0.03 & 0.5 ± 0.02 \\
GCN & KDE-based downsampling & 24.92 ± 1.02 & 0.54 ± 0.03 & 0.52 ± 0.03 & 26.69 ± 2.29 & 0.53 ± 0.02 & 0.51 ± 0.01 \\
RFR & Original & 32.89 ± 1.4 & 0.53 ± 0.05 & 0.53 ± 0.04 & 33.2 ± 1.31 & 0.53 ± 0.05 & 0.53 ± 0.04 \\
RFR & KDE-based reweighting & 26.61 ± 0.88 & 0.54 ± 0.04 & 0.53 ± 0.04 & 29.01 ± 2.06 & 0.55 ± 0.05 & 0.53 ± 0.03 \\
RFR & KDE-guided SMOGN & 39.46 ± 2.99 & 0.06 ± 0.09 & 0.04 ± 0.1 & 35.6 ± 7.39 & 0.33 ± 0.16 & 0.33 ± 0.15 \\
RFR & KDE-based downsampling & 24.39 ± 0.69 & 0.54 ± 0.04 & 0.53 ± 0.03 & 24.61 ± 0.99 & 0.54 ± 0.04 & 0.53 ± 0.04 \\
XGBR & Original & 32.99 ± 1.2 & 0.54 ± 0.04 & 0.53 ± 0.03 & 32.93 ± 1.25 & 0.54 ± 0.04 & 0.53 ± 0.04 \\
XGBR & KDE-based reweighting & 27.98 ± 1.58 & 0.54 ± 0.05 & 0.52 ± 0.04 & 31.7 ± 2.54 & 0.55 ± 0.05 & 0.54 ± 0.04 \\
XGBR & KDE-guided SMOGN & 36.74 ± 1.72 & 0.52 ± 0.04 & 0.52 ± 0.04 & 38.32 ± 1.63 & 0.51 ± 0.05 & 0.52 ± 0.04 \\
XGBR & KDE-based downsampling & 33.55 ± 1.46 & 0.53 ± 0.04 & 0.51 ± 0.04 & 33.97 ± 1.88 & 0.54 ± 0.03 & 0.52 ± 0.03 \\
\bottomrule
\end{tabular}

\end{sidewaystable}

\clearpage

\section{Discussion}\label{sec12}

This study addresses the broader challenge of imbalanced regression, where prediction performance in sparse but critical regions is often more important than overall accuracy. Using hERG as a representative example, we demonstrate how commonly adopted formulations may obscure such region-specific behaviors. Previous hERG studies have primarily focused on IC50 regression or classification tasks; however, IC50 values that cannot be reliably fitted are typically discarded, while classification models provide only coarse toxicity labels. More importantly, hERG liability is well known to be concentration-dependent, indicating that toxicity is not solely determined by chemical structure but also by exposure level. To better reflect this practical consideration, we formulate the task as predicting activity at specific concentrations. This formulation avoids discarding positive or non-fitting samples, enabling a more realistic screening strategy. Beyond task reformulation, this study highlights that addressing imbalance solely at the data level is insufficient. Instead, we show that model selection criteria and evaluation protocols must be aligned with the regions of interest. By emphasizing region-specific performance rather than global metrics, our results demonstrate that meaningful improvements in sparse regions can be achieved even when overall performance appears unchanged.

Before concluding which imbalance strategy performs better, it is essential to clarify how model performance should be evaluated. Metrics computed over the entire response range are often misleading in highly imbalanced regression settings. In this study, approximately 90\% of the samples fall within the $[-20,20)$ region, causing this dense region to dominate total-region metrics such as MAE and MSE. As a result, improvements in sparse but critical regions may be obscured when only global error metrics are considered. Correlation-based metrics introduce additional complexity. While MAE and MSE primarily reflect the influence of dominant regions, Pearson correlation is sensitive to the global distribution of data points. Previous studies have shown that when data exhibit apparent group structures, the overall Pearson correlation is not simply a weighted average of within-group correlations; instead, it can be strongly affected by inter-group differences and extreme values. Consequently, raw data distributions, including outliers, may either exaggerate or deteriorate Pearson correlation. In contrast, Spearman correlation, which is based on rank ordering, is less sensitive to such distributional effects. For this reason, Pearson and Spearman correlations are jointly reported to assess the consistency of correlation trends. To address these issues, we partition the response space into multiple bins and evaluate model performance within each region separately. This region-wise evaluation enables us to focus on the $[-100,-20)$ interval, which represents the primary region of interest in this study. It provides a more faithful assessment of model behavior in sparse yet practically relevant regimes.

Beyond evaluation, we further align model selection with this region-wise goal by using the validation MSE in the $[-100,-20)$ region as the selection criterion. Importantly, this strategy does not modify the training objective of GCN models or the split-quality functions used in tree-based methods. Instead, it affects only the criterion used to select the final model from multiple trained candidates. As such, this approach should be regarded as an explicit alignment between evaluation priorities and model selection decisions, rather than as overfitting or post-hoc optimization. This choice inevitably introduces a trade-off: prioritizing performance in a specific response region may not yield uniform improvements across the entire target distribution. Nevertheless, our empirical results show that performance within the region of interest is consistently improved under this selection criterion, indicating that region-aware model selection constitutes a necessary and principled step for imbalanced regression problems in which certain target ranges carry disproportionate practical importance.

Focusing on the $[-100,-20)$ region, heterogeneous responses to imbalance strategies are observed across GCN, RFR, and XGBR models. Graph-based models exhibit higher sensitivity to resampling-based interventions, whereas tree-based models appear more robust to distributional shifts. This difference may stem from the fact that representation learning in GCNs depends directly on the empirical data distribution. In contrast, tree-based models rely on partitioning rules that are less affected by moderate changes in sample composition. Among the evaluated strategies, SMOGN introduces additional concerns. The original implementation cannot be applied directly in this study because it treats negative response values as zero, which distorts the target distributions. Moreover, the downsampling ratio in the original implementation cannot be manually adjusted. Although we modified these conditions to enable its application, SMOGN remains fundamentally reliant on synthetic sample generation. In the context of experimental bioactivity data, such oversampling may amplify measurement uncertainty rather than the informative signal. This characteristic may explain the observed deterioration in rank-based metrics, suggesting that aggressive synthetic oversampling is not always suitable for sparse yet noisy regions.

The proposed framework is not limited to hERG assays and can be extended to other endpoint prediction tasks. Data imbalance is pervasive in QSAR studies, as truly toxic or biologically meaningful samples often constitute only a small fraction of the dataset. In such settings, the first and most critical step is to explicitly define the regions of interest based on experimental and pharmacological considerations. If a sparse region does not correspond to a practically relevant response range, then the imbalance itself may not pose a meaningful problem. The definition of region boundaries should therefore be guided jointly by biological interpretation and sample availability, as illustrated in this study. Once the regions of interest are defined, appropriate evaluation metrics and model selection criteria must be chosen accordingly. While previous studies have often relied on weighted global metrics to address imbalance, such approaches implicitly assume that importance can be adequately captured by assigning different weights to individual samples. In contrast, this study adopts region-specific evaluation, which directly assesses model behavior within predefined response intervals. This distinction is crucial, as region-wise metrics provide clearer interpretability and avoid confounding effects introduced by global aggregation. Finally, the effectiveness of imbalance-handling strategies is inherently model-dependent. Different model classes respond differently to resampling or reweighting schemes, suggesting that no single imbalance strategy is universally optimal. Instead, imbalance treatment, evaluation metrics, and model selection criteria should be considered jointly as a coordinated design choice rather than as independent components.

Although imbalance-aware strategies improve performance in the primary region of interest, they do not uniformly benefit all sparse regions, such as the $[20,100)$ interval. Even within modest improvements, both prediction error and correlation metrics in this region remain suboptimal, indicating substantial room for further improvement. One plausible explanation lies in data quality rather than modeling limitations alone. The hERG activity data used in this study are aggregated from multiple experimental sources and primarily serve as an initial high-throughput screening resource. While such data enable large-scale modeling, they inevitably introduce experimental variability and measurement uncertainty. In regions corresponding to weaker or concentration-dependent responses, this uncertainty may dominate the signal, thereby limiting the effectiveness of imbalance-handling strategies. As a result, the absence of consistent improvement in certain sparse regions should not be interpreted as methodological failure, but rather as a reflection of intrinsic data noise and heterogeneity.

\section{Conclusion}\label{sec13}

In this study, we systematically evaluated various imbalance-handling strategies for QSAR regression using both GCN and fingerprint-based models. Our results show that models selected based on the interest region consistently outperform those chosen based on total-region metrics. Notably, the effectiveness of a strategy depends on the model type and the target region considered. These findings underscore the importance of region-aware model evaluation for imbalanced regression tasks. Future work could explore automated methods for defining interest regions and extend this framework to other toxicological endpoints.

\backmatter


\bmhead{List of abbreviations}

\begin{description}

\item[QSAR] Quantitative structure--activity relationship

\item[KDE] Kernel density estimation

\item[hERG] Ether-\textit{à-go-go}-Related Gene

\item[SMOGN] Synthetic Minority Oversampling Technique for Regression with Gaussian Noise

\item[CRC] Concentration--response curve

\item[RF] Random Forest

\item[XGB] Extreme Gradient Boosting

\item[GCN] Graph Convolutional Network

\item[HDI] Highest Density Interval

\item[MSE] Mean squared error

\end{description}

\section*{Declarations}

\subsection*{Availability of data and materials}
The human hERG channel activity dataset used in this study was obtained from the PubChem BioAssay database (AID 588834) \cite{AID588834QHTS}. The source code for the imbalance handling methods and machine learning training pipeline developed in this study is publicly available at \href{https://github.com/sungmingyen/qsar-imbalance-regression}{GitHub repository}.

\subsection*{Competing interests}
The authors declare no conflict of interest.

\subsection*{Funding}
This work was financially supported by National Science and Technology Council (NSTC 114-2221-E-002-167-MY3).

\subsection*{Authors' contributions}
Ming-Yen Sung, Hsin Liu, and Yufeng J. Tseng conceived the study. 
Hsin Liu prepared the dataset. Cheng-Wei Lee optimized the GCN model based on DeepChem. Ming-Yen Sung applied imbalance regression approaches and conducted both fingerprint-based machine learning models and graph-based models in this study. Ming-Yen Sung and Hsin Liu wrote the manuscript. Tung-Sheng Cho and Yufeng J. Tseng supervised the research and provided critical feedback.

\subsection*{Acknowledgements}
This study was performed using computing resources from the Laboratory of Computational Molecular Design and Metabolomics within the Department of Computer Science and Information Engineering at National Taiwan University.





\bibliography{reference}

@misc{AID588834QHTS,
  title = {{{AID}} 588834 - {{qHTS Assay}} for {{Small Molecule Inhibitors}} of the {{Human hERG Channel Activity}} - {{PubChem}}},
  urldate = {2025-08-14},
  howpublished = {https://pubchem.ncbi.nlm.nih.gov/bioassay/588834},
}

@article{al-jubouriImbalanceRegressionApproach2025,
  title = {An {{Imbalance Regression Approach}} to {{Toxicity Prediction}} of {{Chemicals}} for {{Potential Use}} in {{Environmentally Acceptable Lubricants}}},
  author = {{Al-Jubouri}, B. and Desiati, I. and Wijanarko, W. and Espallargas, N.},
  year = 2025,
  month = mar,
  journal = {ACS Applied Materials \& Interfaces},
  volume = {17},
  number = {11},
  pages = {16725--16737},
  publisher = {American Chemical Society},
  issn = {1944-8244},
  doi = {10.1021/acsami.4c10622},
  urldate = {2025-10-17},
  
}

@inproceedings{brancoSMOGNPreprocessingApproach2017,
  title = {{{SMOGN}}: A {{Pre-processing Approach}} for {{Imbalanced Regression}}},
  shorttitle = {{{SMOGN}}},
  booktitle = {Proceedings of the {{First International Workshop}} on {{Learning}} with {{Imbalanced Domains}}: {{Theory}} and {{Applications}}},
  author = {Branco, Paula and Torgo, Lu{\'i}s and Ribeiro, Rita P.},
  year = 2017,
  month = oct,
  pages = {36--50},
  publisher = {PMLR},
  issn = {2640-3498},
  urldate = {2025-10-17},
  langid = {english},
  
}

@misc{FluxORPotassiumIon,
  title = {{{FluxOR Potassium Ion Channel Assay}} - {{TW}}},
  urldate = {2025-08-19},
  howpublished = {https://www.thermofisher.com/tw/zt/home/industrial/pharma-biopharma/drug-discovery-development/target-and-lead-identification-and-validation/ion-channel-biology/cell-based-ion-channel-assays/fluorescent-fluxor-trade-.html},
  langid = {english},
}

@article{kaczorowskiImportanceBeingProfiled2011,
  title = {The {{Importance}} of {{Being Profiled}}: {{Improving Drug Candidate Safety}} and {{Efficacy Using Ion Channel Profiling}}},
  shorttitle = {The {{Importance}} of {{Being Profiled}}},
  author = {Kaczorowski, Gregory J. and Garcia, Maria L. and Bode, Jacob and Hess, Stephen D. and Patel, Umesh A.},
  year = 2011,
  month = dec,
  journal = {Frontiers in Pharmacology},
  volume = {2},
  publisher = {Frontiers},
  issn = {1663-9812},
  doi = {10.3389/fphar.2011.00078},
  urldate = {2026-01-06},
  langid = {english},
  
}

@article{krishnaHighThroughputChemicalScreening2022,
  title = {High-{{Throughput Chemical Screening}} and {{Structure-Based Models}} to {{Predict hERG Inhibition}}},
  author = {Krishna, Shagun and Borrel, Alexandre and Huang, Ruili and Zhao, Jinghua and Xia, Menghang and Kleinstreuer, Nicole},
  year = 2022,
  month = feb,
  journal = {Biology},
  volume = {11},
  number = {2},
  pages = {209},
  publisher = {Multidisciplinary Digital Publishing Institute},
  issn = {2079-7737},
  doi = {10.3390/biology11020209},
  urldate = {2025-08-19},
  copyright = {http://creativecommons.org/licenses/by/3.0/},
  langid = {english},
}

@article{liFingerprintingInteractionsProteins2024,
  title = {Fingerprinting {{Interactions}} between {{Proteins}} and {{Ligands}} for {{Facilitating Machine Learning}} in {{Drug Discovery}}},
  author = {Li, Zoe and Huang, Ruili and Xia, Menghang and Patterson, Tucker A. and Hong, Huixiao},
  year = 2024,
  month = jan,
  journal = {Biomolecules},
  volume = {14},
  number = {1},
  pages = {72},
  issn = {2218-273X},
  doi = {10.3390/biom14010072},
  langid = {english},
  pmcid = {PMC10813698},
  pmid = {38254672},
}

@inproceedings{liImbalancedToxicityPrediction2020,
  title = {Imbalanced {{Toxicity Prediction Using Multi-Task Learning}} and {{Over-Sampling}}},
  booktitle = {2020 {{International Conference}} on {{Machine Learning}} and {{Cybernetics}} ({{ICMLC}})},
  author = {LI, JIN-CHENG},
  year = 2020,
  month = feb,
  pages = {1--7},
  issn = {2160-1348},
  doi = {10.1109/ICMLC51923.2020.9469546},
  urldate = {2026-01-06},
}

@article{liReviewQuantitativeStructureactivity2025,
  title = {A Review of Quantitative Structure-Activity Relationship: {{The}} Development and Current Status of Data Sets, Molecular Descriptors and Mathematical Models},
  shorttitle = {A Review of Quantitative Structure-Activity Relationship},
  author = {Li, Jianmin and Zhao, Tian and Yang, Qin and Du, Shijie and Xu, Lu},
  year = 2025,
  month = jan,
  journal = {Chemometrics and Intelligent Laboratory Systems},
  volume = {256},
  pages = {105278},
  issn = {0169-7439},
  doi = {10.1016/j.chemolab.2024.105278},
  urldate = {2026-01-06},
}

@article{lovricPredictiveCapabilityQSAR2021,
  title = {Predictive {{Capability}} of {{QSAR Models Based}} on the {{CompTox Zebrafish Embryo Assays}}: {{An Imbalanced Classification Problem}}},
  shorttitle = {Predictive {{Capability}} of {{QSAR Models Based}} on the {{CompTox Zebrafish Embryo Assays}}},
  author = {Lovri{\'c}, Mario and Malev, Olga and Klobu{\v c}ar, G{\"o}ran and Kern, Roman and Liu, Jay J. and Lu{\v c}i{\'c}, Bono and Lovri{\'c}, Mario and Malev, Olga and Klobu{\v c}ar, G{\"o}ran and Kern, Roman and Liu, Jay J. and Lu{\v c}i{\'c}, Bono},
  year = 2021,
  month = mar,
  journal = {Molecules},
  volume = {26},
  number = {6},
  publisher = {publisher},
  issn = {1420-3049},
  doi = {10.3390/molecules26061617},
  urldate = {2026-01-06},
  copyright = {http://creativecommons.org/licenses/by/3.0/},
  langid = {english},
}

@article{ribeiroImbalancedRegressionExtreme2020,
  title = {Imbalanced Regression and Extreme Value Prediction},
  author = {Ribeiro, Rita P. and Moniz, Nuno},
  year = 2020,
  month = sep,
  journal = {Machine Learning},
  volume = {109},
  number = {9},
  pages = {1803--1835},
  issn = {1573-0565},
  doi = {10.1007/s10994-020-05900-9},
  urldate = {2026-01-02},
  langid = {english},
}

@article{rogersExtendedConnectivityFingerprints2010,
  title = {Extended-{{Connectivity Fingerprints}}},
  author = {Rogers, David and Hahn, Mathew},
  year = 2010,
  month = may,
  journal = {Journal of Chemical Information and Modeling},
  volume = {50},
  number = {5},
  pages = {742--754},
  publisher = {American Chemical Society (ACS)},
  issn = {1549-9596, 1549-960X},
  doi = {10.1021/ci100050t},
  urldate = {2025-07-22},
  langid = {english},
}

@article{steiningerDensitybasedWeightingImbalanced2021,
  title = {Density-Based Weighting for Imbalanced Regression},
  author = {Steininger, Michael and Kobs, Konstantin and Davidson, Padraig and Krause, Anna and Hotho, Andreas},
  year = 2021,
  month = aug,
  journal = {Machine Learning},
  volume = {110},
  number = {8},
  pages = {2187--2211},
  issn = {1573-0565},
  doi = {10.1007/s10994-021-06023-5},
  urldate = {2025-09-30},
  langid = {english},
}

@article{vasilevComprehensiveReviewApplication2025,
  title = {A ({{Comprehensive}}) {{Review}} of the {{Application}} of {{Quantitative Structure}}--{{Activity Relationship}} ({{QSAR}}) in the {{Prediction}} of {{New Compounds}} with {{Anti-Breast Cancer Activity}}},
  author = {Vasilev, Boris and Atanasova, Mariyana},
  year = 2025,
  month = jan,
  journal = {Applied Sciences},
  volume = {15},
  number = {3},
  pages = {1206},
  publisher = {Multidisciplinary Digital Publishing Institute},
  issn = {2076-3417},
  doi = {10.3390/app15031206},
  urldate = {2025-08-20},
  copyright = {http://creativecommons.org/licenses/by/3.0/},
  langid = {english},
}

@misc{yangDelvingDeepImbalanced2021,
  title = {Delving into {{Deep Imbalanced Regression}}},
  author = {Yang, Yuzhe and Zha, Kaiwen and Chen, Ying-Cong and Wang, Hao and Katabi, Dina},
  year = 2021,
  month = may,
  number = {arXiv:2102.09554},
  eprint = {2102.09554},
  primaryclass = {cs},
  publisher = {arXiv},
  doi = {10.48550/arXiv.2102.09554},
  urldate = {2025-09-11},
  archiveprefix = {arXiv},
}

@article{zakharovQSARModelingImbalanced2014,
  title = {{{QSAR Modeling}} of {{Imbalanced High-Throughput Screening Data}} in {{PubChem}}},
  author = {Zakharov, AlexeyV. and Peach, Megan L. and Sitzmann, Markus and Nicklaus, Marc C.},
  year = 2014,
  month = mar,
  journal = {Journal of Chemical Information and Modeling},
  volume = {54},
  number = {3},
  pages = {705--712},
  issn = {1549-9596},
  doi = {10.1021/ci400737s},
  urldate = {2025-10-17},
  pmcid = {PMC3985743},
  pmid = {24524735},
}

@article{zhangNovelGradientBoosting2024,
  title = {A Novel Gradient Boosting Approach for Imbalanced Regression},
  author = {Zhang, Wenchao and Shi, Peixin and Jia, Pengjiao and Zhou, Xiaoqi},
  year = 2024,
  month = oct,
  journal = {Neurocomputing},
  volume = {601},
  pages = {128091},
  issn = {0925-2312},
  doi = {10.1016/j.neucom.2024.128091},
  urldate = {2025-09-18},
 
}

@misc{deepchemGCNTutorial,
  title = {Introduction to Graph Convolutions},
  author = {{DeepChem Contributors}},
  year = {2025},
  howpublished = {\url{https://deepchem.io/tutorials/introduction-to-graph-convolutions/}},
  note = {Accessed: 2026-05-08}
}

\end{document}